\documentclass[10pt,conference]{IEEEtran}
\IEEEoverridecommandlockouts
\usepackage{cite}
\usepackage{amsmath,amssymb,amsfonts}
\usepackage{algorithmic}
\usepackage{graphicx}
\usepackage{textcomp}
\usepackage{xcolor}

\usepackage{times}
\usepackage{latexsym}
\usepackage{multirow}
\usepackage{soul}
\usepackage{array}
\usepackage{booktabs}
\usepackage{subcaption}
\usepackage{hyperref}
\usepackage{tikz}
\usetikzlibrary{shapes,arrows,positioning,fit,backgrounds}

\makeatletter
\newcommand{\linebreakand}{%
  \end{@IEEEauthorhalign}
  \hfill\mbox{}\par
  \mbox{}\hfill\begin{@IEEEauthorhalign}
}
\makeatother

\begin{document}

\title{Position Paper: Just as Humans Need Vaccines, So Do Models - Immunization to Combat Falsehoods}

\author{
    \IEEEauthorblockN{Shaina Raza\textsuperscript{*}\thanks{\textsuperscript{*}Corresponding author. Email: shaina.raza@torontomu.ca}}
    \IEEEauthorblockA{\textit{Vector Institute} \\
    Toronto, Canada}
    \and
    \IEEEauthorblockN{Rizwan Qureshi}
    \IEEEauthorblockA{\textit{University of Central Florida} \\
    Orlando, USA}
    \and
    \IEEEauthorblockN{Azib Farooq}
    \IEEEauthorblockA{\textit{University of Cincinnati} \\
    Cincinnati, USA}
    \and
    \IEEEauthorblockN{Marcelo Lotif}
    \IEEEauthorblockA{\textit{Vector Institute} \\
    Toronto, Canada}

    \linebreakand

    \IEEEauthorblockN{Aman Chadha\textsuperscript{$\dagger$}\thanks{\textsuperscript{$\dagger$}Work done outside of Amazon.}}
    \IEEEauthorblockA{\textit{Independent Researcher} \\
    USA}
    \and
    \IEEEauthorblockN{Deval Pandya}
    \IEEEauthorblockA{\textit{Vector Institute} \\
    Toronto, Canada}
    \and
    \IEEEauthorblockN{Christos Emmanouilidis}
    \IEEEauthorblockA{\textit{University of Groningen} \\
    The Netherlands}
}
\maketitle

\begin{abstract}
Large language models (LLMs) reproduce misinformation by learning the linguistic patterns that make falsehoods persuasive, ssuch as hedging, false presuppositions, and citation fabrication, rather than merely memorizing false facts. We propose model immunization: supervised fine-tuning on curated (false claim, correction) pairs injected as small "vaccine doses" (5–10\% of tokens) alongside truthful data. Unlike post-hoc filtering or preference-based alignment, immunization provides direct negative supervision on labeled falsehoods. Across four open-weight model families, immunization improves TruthfulQA accuracy by 12 points and misinformation rejection by 30 points with negligible capability loss. We outline design requirements, which includes, dosage, labeling, quarantine, diversity and call for standardized vaccine corpora and benchmarks that test generalization, making immunization a routine component of responsible LLM development. The project webpage is available at \href{https://shainarazavi.github.io/ai-vaccine/}{\textbf{project page}} .
\end{abstract}


 \section{Introduction}
When a language model asserts that ``vaccines cause autism''~\cite{wardle2017information} or that ``the 2020 U.S.\ election was stolen''~\cite{rashkin2017truth}, it is not simply retrieving a memorized false fact. It has learned something deeper: the linguistic conventions through which such claims are constructed and made persuasive. These include hedging patterns (``some people say...''), pre-suppositional triggers that smuggle in false premises, citation fabrications that mimic authoritative sourcing, and rhetorical structures that lend credibility to baseless claims. The problem, in other words, is not merely one of factual knowledge but of linguistic competence, which means the models have learned \textit{how} misinformation sounds, and they reproduce those patterns regardless of truth value.

\begin{figure}[t]
  \centering
  \begin{subfigure}{\columnwidth}
    \centering
    \includegraphics[width=0.8\linewidth]{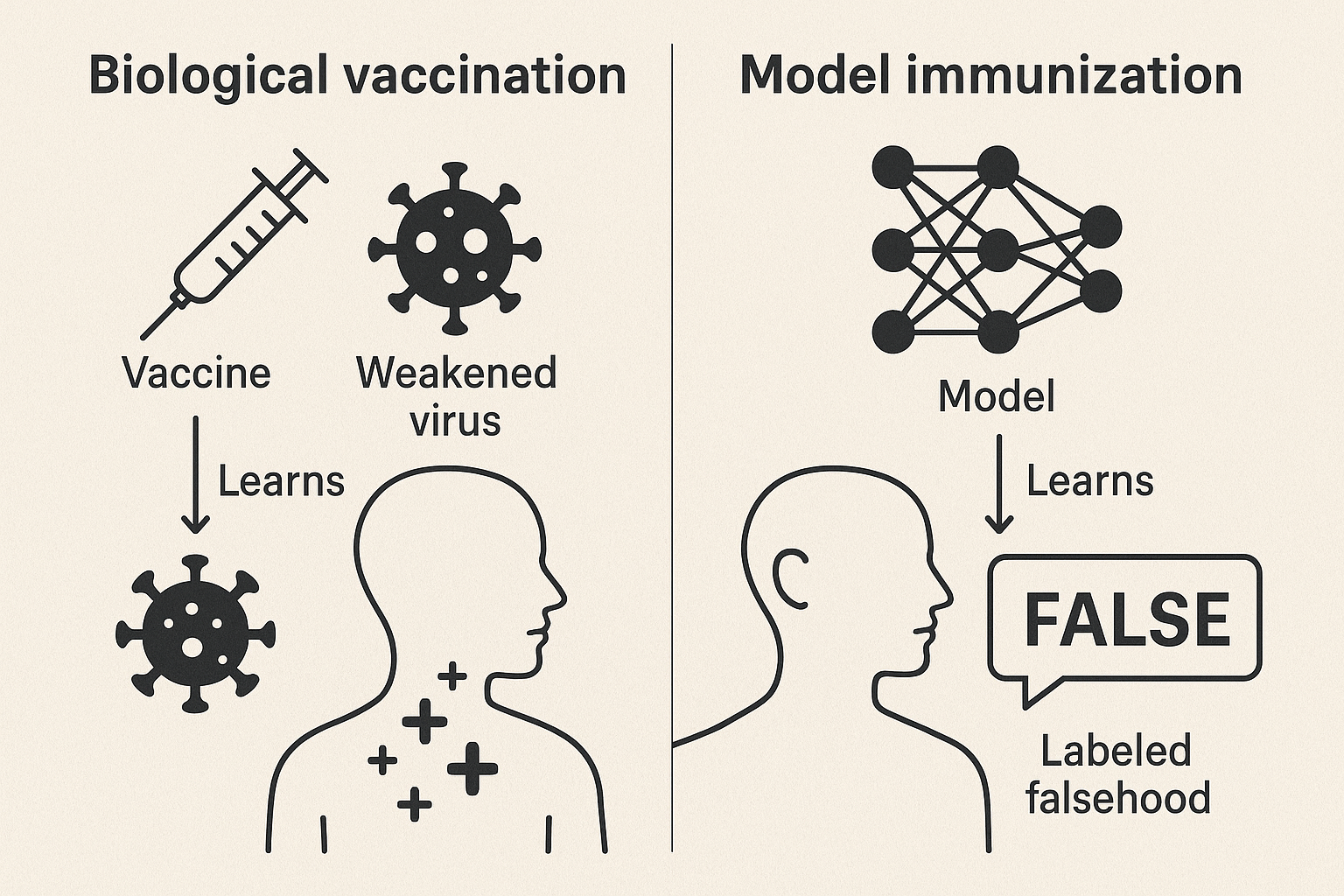}
    \caption{Biological vaccination vs.\ model immunization: controlled exposure builds resistance.}
    \label{fig:vaccine-analogy}
  \end{subfigure}

  \begin{subfigure}{\columnwidth}
    \centering
    \includegraphics[width=0.8\linewidth]{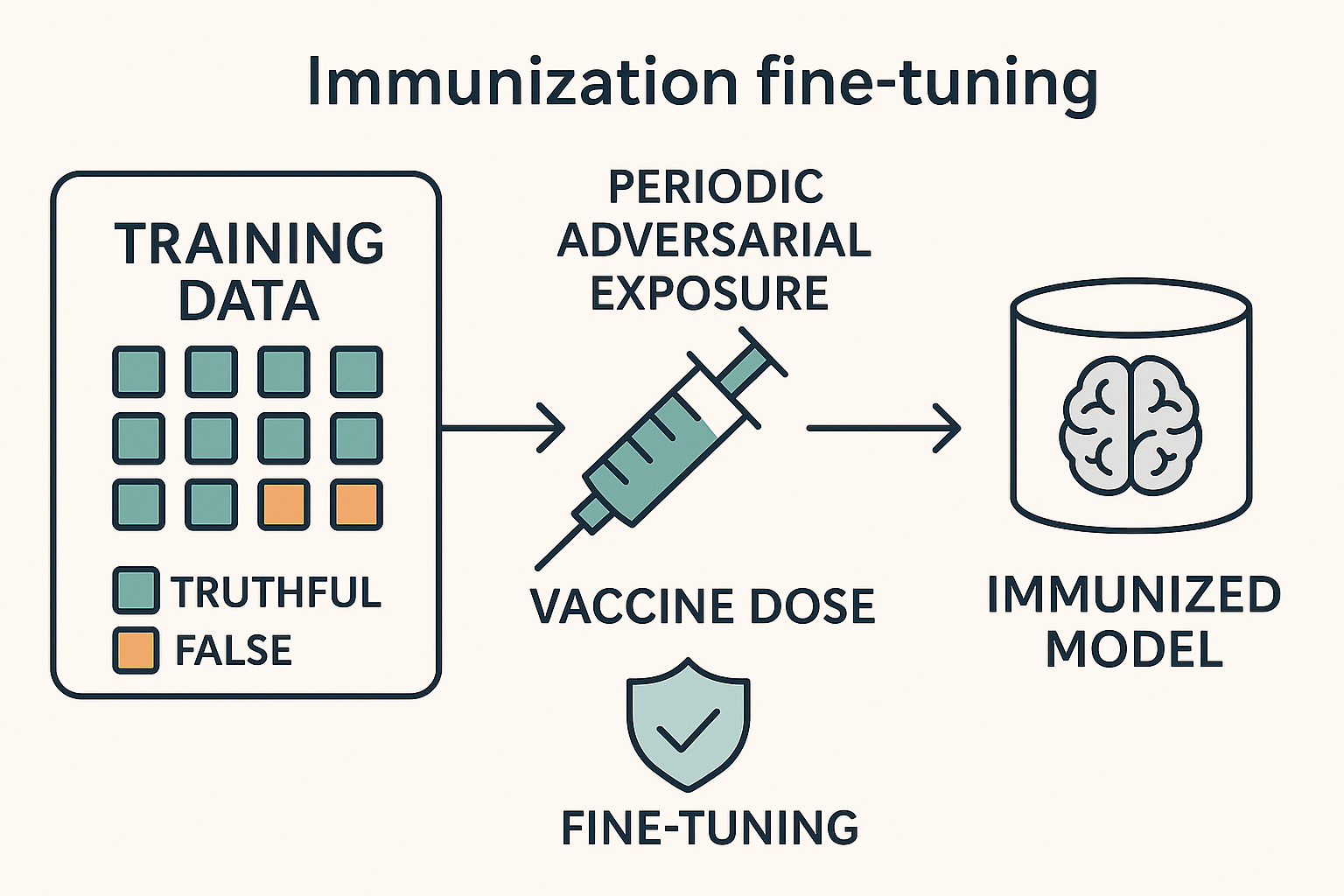}
    \caption{Immunization fine-tuning: periodic injection of labeled falsehoods (5–10\%) alongside truthful data.}
    \label{fig:imm}
  \end{subfigure}
  \caption{The immunization analogy. Just as biological vaccines use controlled exposure to weakened pathogens to train immune responses, model immunization uses controlled exposure to labeled falsehoods to train rejection responses.}
  \label{fig:analogy}
  \vspace{-1em}
\end{figure}

Current approaches to improving model truthfulness do not directly address this phenomenon. Post-hoc detection systems attempt to catch misinformation after generation~\cite{zhou2024correcting}; for example, they treat symptoms rather than causes. Reinforcement learning from human feedback (RLHF) provides indirect pressure toward truthfulness via reward signals~\cite{ouyang2022training}, but does not give models explicit supervision on what makes a claim false; therefore, truthfulness is diluted across helpfulness, harmlessness, and stylistic preferences. Adversarial training harden models against input perturbations~\cite{madry2017towards}, such as typos, paraphrases, adversarial suffixes, but does not address the semantic content of misinformation. A model robust to character-level attacks may still enthusiastically endorse false claims presented in clean text.

We propose a more direct intervention: \textbf{supervised immunization against misinformation}. The core idea is simple but represents a conceptual shift grounded in psychological inoculation theory~\cite{mcguire1961resistance}, which demonstrates that controlled exposure to weakened forms of persuasive attacks can build cognitive resistance in humans~\cite{roozenbeek2022psychological}. Rather than filtering false data out of training corpora, we deliberately include it, but with explicit negative labels and paired corrections. Just as \textbf{biological vaccines} use controlled exposure to weakened pathogens to train immune responses (Figure~\ref{fig:vaccine-analogy}), \textbf{model immunization} uses controlled exposure to labeled falsehoods to train rejection responses.
\begin{figure*}[t]
    \centering
    \includegraphics[width=0.8\textwidth]{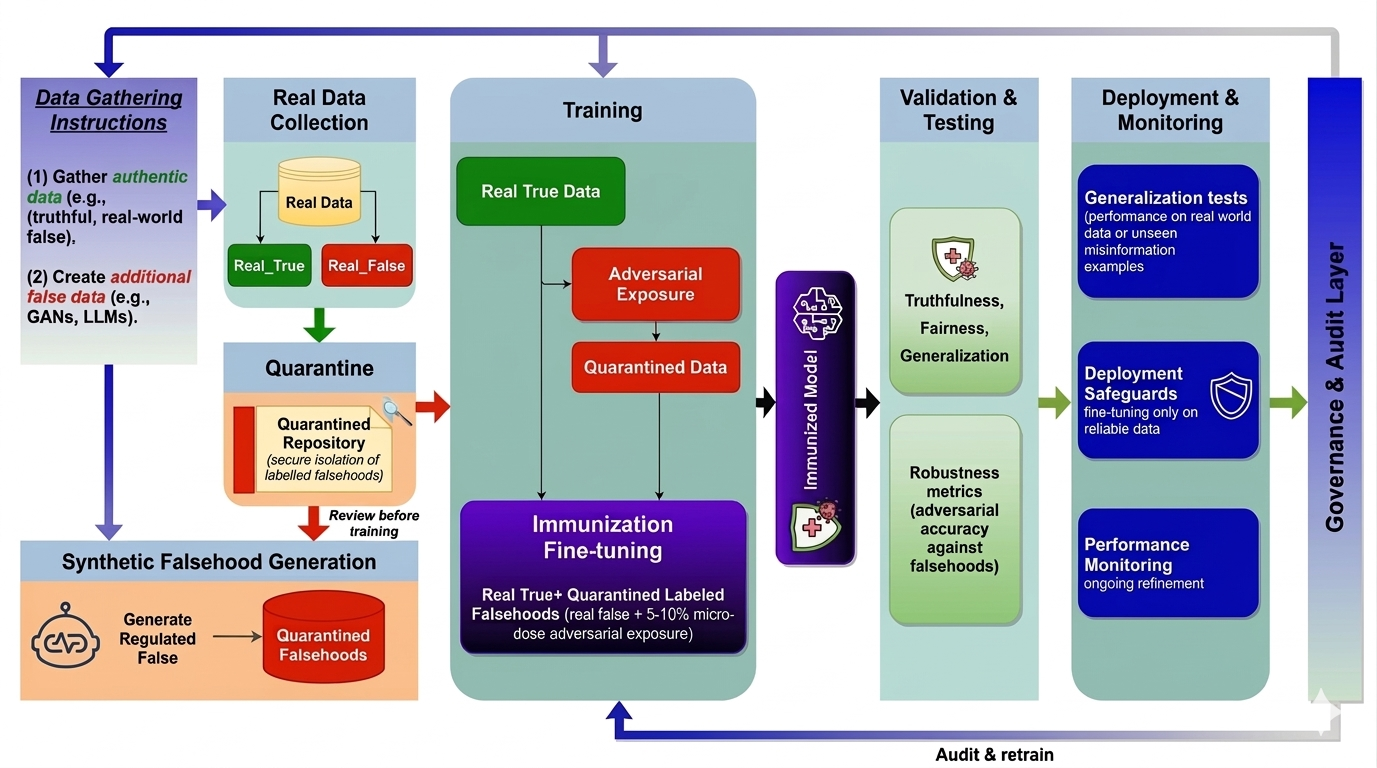}
    \caption{The immunization pipeline. Data curation separates truthful and false examples; falsehoods enter a quarantined repository with fact-checker verification before controlled injection during fine-tuning. Validation tests truthfulness and robustness to held-out misinformation; deployment includes continuous monitoring with feedback for iterative refinement. A governance layer ensures accountability and auditability throughout.}
    \label{fig:pipeline}
\end{figure*}
In practice, this means periodically injecting small ``doses'' of fact-checked false claims (5–10\% of fine-tuning tokens) alongside truthful data (Figure~\ref{fig:imm}), with each falsehood explicitly marked as false and paired with correct information. This reframes the role of false data in alignment. Standard practice treats misinformation as contamination; we argue it should be treated as \textit{evasive signal}, information if properly labeled and governed, teaches models what to avoid. The approach is also analogous to how toxic language datasets are used to train content filters~\cite{gehman2020realtoxicityprompts}: we do not hide offensive content from the model but show it explicitly, with labels, so the model learns to recognize and reject it.

Our \textbf{position} rests on three claims. First, \textit{truthfulness requires negative supervision}. Preference-based methods like RLHF encode factuality as one objective among many; models need explicit examples of falsehoods labeled as such: direct negative signal, not indirect preference pressure. Second, \textit{misinformation is fundamentally a linguistic phenomenon}. False claims exhibit characteristic patterns that models learn and reproduce; immunization should target these patterns, not just specific false propositions. Third, \textit{the AI and NLP communities need new infrastructure} to make immunization practical: standardized vaccine corpora, benchmarks testing robustness to unseen misinformation, and evaluation protocols measuring generalization across languages and domains.

\section{The Linguistic Case for Misinformation Research}
\label{sec:linguistic}

Misinformation is fundamentally a \emph{linguistic} phenomenon. Beyond incorrect facts, false claims exhibit recurring discourse patterns that make them persuasive, including hedged assertions, false presuppositions, citation fabrication, emotive amplification, and false balance. These patterns recur across domains such as health, politics, and science, and are learned by language models from web-scale corpora.
Table~\ref{tab:linguistic-patterns} summarizes six such linguistic signatures commonly observed in misinformation. Crucially, these patterns generalize across claims: removing specific false statements from training data does not remove the underlying discourse templates. As a result, models may correctly answer factual queries while still reproducing misinformation when prompted with persuasive linguistic framing.
This observation motivates training interventions that target \emph{patterns}, not just propositions. Improving truthfulness therefore requires teaching models to recognize and reject misinformation-associated linguistic features, rather than relying solely on fact recall or post-hoc filtering.

\begin{table}[t]
\centering
\renewcommand{\arraystretch}{0.85} 
\footnotesize
\caption{Linguistic patterns characteristic of misinformation. Each represents a learnable feature that models acquire from training data.}
\begin{tabular}{p{2cm}p{5.5cm}}
\toprule
\textbf{Pattern} & \textbf{Description and Example} \\
\midrule
Hedged assertion & Weasel words avoiding commitment: ``Some experts believe...'', ``It has been reported that...'' \\
\addlinespace
False presupposition & Smuggles false premise into discourse: ``When did the government admit the cover-up?'' \\
\addlinespace
Citation fabrication & Mimics scholarly sourcing: ``According to a Stanford study...'' (nonexistent) \\
\addlinespace
Emotive amplification & Substitutes affect for evidence: ``The SHOCKING truth they don't want you to know'' \\
\addlinespace
False balance & Presents fringe views as equally credible: ``While most scientists say X, others argue Y'' \\
\addlinespace
Temporal manipulation & Implies causation through proximity: ``Shortly after the vaccine rollout, deaths increased'' \\
\bottomrule
\end{tabular}
\label{tab:linguistic-patterns}
\end{table}

\section{Immunization as Negative Supervision}
\label{sec:immunization}
Having established that misinformation is a linguistic phenomenon, we now present model immunization as a direct intervention targeting the patterns described above.
\subsection{The Core Proposal}
Model immunization is supervised fine-tuning on a curated corpus of (false claim, correction) pairs, where each false claim carries an explicit negative label and the model learns to reject it or generate the correction rather than endorse the claim. The approach treats fact-checked falsehoods not as contamination to be removed but as training signal to be harnessed, which is analogous to how exposure to weakened pathogens trains biological immune systems. The framework requires \textbf{four} design choices. 

\textbf{(1)} \textit{Dosage} determines what fraction of fine-tuning tokens come from the vaccine corpus. Too little yields no measurable effect; too much risks teaching the model to generate misinformation rather than reject it. Preliminary experiments suggest 5–10\% strikes an appropriate balance, though the optimal dose likely varies with model scale, domain, and the diversity of the vaccine corpus. 

\textbf{(2)} \textit{Labeling} ensures the model learns falsehoods as negative examples rather than internalizing them as knowledge. Each claim is paired with an explicit ``this is false'' marker and a correction providing the factual alternative. The training objective penalizes the model for reproducing or endorsing the false claim and rewards it for generating the correction or refusing to engage. Without clear labeling, the distinction between vaccine data and knowledge data collapses, and immunization becomes poison.

\textbf{(3)} \textit{Quarantine} keeps vaccine data strictly separate from the model's knowledge base throughout the training pipeline. Falsehoods must never be retrievable as facts; they exist only as patterns to reject. This separation is both technical, distinct data pipelines, separate storage, clear provenance tracking, and governance-related, with audit trails documenting every falsehood's source, verification, and use in training. 

\textbf{(4)} \textit{Diversity} requires coverage across misinformation types and linguistic patterns, not just specific claims. A vaccine corpus containing only health misinformation will not immunize against political falsehoods; one containing only English claims will not transfer to other languages. Effective immunization demands breadth across domains (health, politics, science, history), across languages, and across the linguistic signatures described in Section~\ref{sec:linguistic}. The goal is pattern-level immunity, not instance-level memorization.

\begin{figure*}[h]
\centering

\begin{subfigure}[t]{0.49\textwidth}
  \centering
  \vspace{0pt} 
  
  \scriptsize
  \setlength{\tabcolsep}{2.5pt}
  \renewcommand{\arraystretch}{1.05}
  \caption{Comparison by input type, goal, and factuality signal.}
  \begin{tabular}{@{}p{2.05cm}p{1.55cm}p{1.55cm}p{1.55cm}@{}}
    \toprule
    \textbf{Technique} & \textbf{Input} & \textbf{Goal} & \textbf{Signal} \\
    \midrule
    Adversarial training & perturbations & robustness & none \\
    RLHF & preferences & alignment & indirect \\
    \textbf{Immunization} & \textbf{falsehoods} & \textbf{truthfulness} & \textbf{direct} \\
    Post-hoc detection & outputs & filtering & reactive \\
    \bottomrule
  \end{tabular}
  
  \label{fig:defense-overview-table}
\end{subfigure}
\hfill
\begin{subfigure}[t]{0.49\textwidth}
  \centering
  \vspace{0pt}

  \resizebox{\linewidth}{!}{%
    \begin{tikzpicture}[stage/.style={font=\footnotesize, above},
      box/.style={rectangle,rounded corners=3pt,minimum height=0.7cm,minimum width=1.8cm,
      font=\footnotesize\bfseries,text=white,align=center},
      timeline/.style={thick,gray}]
      \draw[timeline] (0,0) -- (11,0);
      \foreach \x in {0,3.5,7.5,11}{\draw[timeline] (\x,0.15) -- (\x,-0.15);}
      \node[stage] at (1.75,0.25) {Training};
      \node[stage] at (5.5,0.25) {Fine-tuning};
      \node[stage] at (9.25,0.25) {Inference};
      \node[box, fill=blue!60]    at (1.75,-0.8) {Adversarial\\[-2pt]Training};
      \node[box, fill=teal!70]    at (4.5,-0.8)  {RLHF};
      \node[box, fill=violet!70] at (6.75,-0.8) {Model\\[-2pt]Immunization};
      \node[box, fill=orange!80] at (9.25,-0.8) {Post-hoc\\[-2pt]Detection};
    \end{tikzpicture}
  }
  \caption{Lifecycle timeline: when each defense applies.}
  \label{fig:defense-overview-timeline}
\end{subfigure}

\caption{Misinformation defense techniques across LLM lifecycle. Immunization provides direct negative supervision on labeled falsehoods during fine-tuning, distinct from indirect preference signals (RLHF) and reactive output filtering (post-hoc detection).}
\label{fig:defense-overview}
\end{figure*}

Figure~\ref{fig:pipeline} illustrates the complete pipeline. The process begins with data curation, where authentic data and real-world falsehoods are collected from reliable sources, for example, fact-checking organizations, debunking databases, and known hoax repositories. These are augmented with synthetic falsehoods generated to fill coverage gaps, ensuring the vaccine corpus spans the full space of misinformation patterns. All false claims enter a quarantined repository where they are reviewed, labeled, and paired with corrections before any training use.

During immunization fine-tuning, the model receives a controlled mixture of standard training data and vaccine doses. The quarantined falsehoods are injected periodically, not concentrated in a single phase but distributed across training to reinforce rejection patterns throughout. Validation testing then evaluates both truthfulness (\textit{does the model reject known falsehoods?}) and generalization \textit{(does it reject \textit{novel} falsehoods exhibiting similar patterns?}). Deployment includes ongoing monitoring for emerging misinformation and mechanisms for ``\textbf{booster}'' updates when new false narratives appear.
Unlike standard supervised setups, immunization fine-tuning exposes the model only to\textbf{ false claims with fact-checker corrections}, it is equivalent to training response behavior rather than true–false classification.

\subsection{Distinguishing Immunization from Related Approaches}

\textbf{Model immunization} occupies a distinct niche in the landscape of misinformation defenses. Figure~\ref{fig:defense-overview} positions it relative to adversarial training, RLHF, and post-hoc detection along two dimensions: when in the model lifecycle the intervention occurs, and what type of signal it provides for factuality. 
The comparison with \textbf{RLHF} is particularly important, where human raters  prefer truthful responses, but this signal is diluted across helpfulness, and other criteria, where truthfulness competes with multiple objectives \cite{ouyang2022training}. Immunization provides \textit{direct} supervision: each false claim is explicitly labeled and paired with a correction. 

\textbf{Adversarial training} targets a different problem: robustness to input perturbations like typos and adversarial suffixes \cite{madry2017towards}. This hardens models against \textit{how} inputs are presented, not \textit{what} they claim. A model resistant to typo-based attacks may still endorse misinformation in clean prose. The two approaches are complementary.
\textbf{Data decontamination}, such as filtering falsehoods from training corpora, removes specific claims but does not teach rejection as a skill \cite{dodge2021documenting}. A decontaminated model has never seen certain falsehoods; it has not learned to refuse them. Immunization inverts this logic: include falsehoods with explicit negative labels so the model learns what to reject.
\textbf{Post-hoc detection} catches misinformation after generation \cite{guo2022survey}. This is reactive, as errors must first be produced, then filtered. Immunization is preventive that reduces the burden on downstream classifiers by addressing the root cause.

\subsection{Governance Requirements}
Using false data as a training signal raises legitimate concerns that require structured governance. Without safeguards,``immunization'' could become a vector for data poisoning, deliberately teaching models misinformation under the guise of inoculation. We address this through five governance principles. First, \textit{transparency} requires documenting every use of false data in training, ensuring each quarantined falsehood is traceable from source to final model with audit logs recording provenance and labeling decisions \cite{chen2025transparent}. Second, \textit{no promotion of false content} ensures every curated false statement is treated as a negative training signal, with the training objective penalizing reproduction or endorsement of false claims while rewarding corrections or refusals~\cite{raza2022fake}. Third, \textit{alignment with human values} focuses curation on clearly discredited falsehoods with broad consensus such as dangerous health misinformation, debunked conspiracy theories, and verified hoaxes, while contested claims require additional human oversight \cite{larsen2024aivaluealignment}. Fourth, \textit{preventing abuse} distinguishes responsible immunization from covert poisoning through open documentation, shared protocols, and use of publicly verifiable fact-checked datasets. Finally, \textit{continuous accountability} establishes mechanisms for ongoing oversight, including public reporting channels that allow users to flag cases where immunization may have failed, feeding back into corpus refinement and model updates \cite{busuioc2021accountable}.

\section{A Research Agenda for the AI and NLP Communities}
\label{sec:agenda}
Model immunization cannot succeed without new infrastructure. Current resources are inadequate in three respects: benchmarks measure the wrong capabilities, appropriate training corpora do not exist at scale, and evaluation protocols ignore generalization. This section outlines what the community must build.

\subsection{Benchmarks That Test Resistance, Not Recall}

Existing factuality benchmarks test whether models \textit{know} true facts, for example, TruthfulQA \cite{lin2022truthfulqa} measures avoidance of false claims, FActScore \cite{min2023factscore} measures factual precision. But knowing facts and resisting misinformation are different capabilities. A model might correctly answer ``Is the earth round?'' while generating flat-earth content when prompted to argue for it, or refuse to state ``Vaccines cause autism'' while hedging when asked about vaccine-autism relationships.
We need benchmarks that test \textbf{defensive} capabilities: \textbf{endorsement resistance} (does the model refuse to validate false claims?), \textbf{elicitation resistance} (does it refuse prompts designed to generate misinformation?), \textbf{presupposition detection} (does it correct false premises rather than accepting them?), and \textbf{cross-type generalization} (does resistance transfer to unseen misinformation categories?). Such a benchmark would complement existing measures by testing defense, not just recall.

\subsection{Vaccine Corpora That Enable Systematic Training}
Ad-hoc falsehood datasets produce fragmentation, for example, variable quality, inconsistent labeling, irreproducible results. The community needs shared resources analogous to FEVER \cite{thorne2018fever}, but designed for negative supervision. Effective vaccine corpora require: \textbf{multilingual coverage} from independent fact-checkers across languages; \textbf{linguistic annotation} tagging patterns from Section~\ref{sec:linguistic} (hedging, presupposition, citation fabrication); \textbf{domain stratification} by topic (health, politics, science) for cross-domain studies; and \textbf{synthetic augmentation} via template-based generation to span possible misinformation beyond observed instances. Building such corpora requires collaboration between AI and NLP researchers, fact-checkers, and domain experts, but without standardized resources, immunization research will remain fragmented.
\subsection{Evaluation Protocols That Demand Generalization}
Demonstrating that a model rejects the specific falsehoods it was trained on proves little. Memorization is easy; generalization is hard~\cite{volpi2018generalizing}. If immunization only teaches models to reject the 500 claims in the vaccine corpus, it offers no protection against the boundless space of misinformation in the wild.
Evaluation protocols must therefore require generalization tests as standard, not optional. \textbf{Held-out claim types} test whether immunization against one category (e.g., health misinformation) transfers to another (e.g., political misinformation) absent from training. Evaluation protocols must require generalization tests as standard, not optional. Without testing transfer across domains, linguistic framings, languages, and time, immunization risks overfitting to specific false claims rather than learning robust resistance patterns.
\textbf{Temporal generalization} tests whether immunization against established misinformation helps with newly emerging false claims. Without these generalization requirements, immunization research risks producing a literature of overfitted results that collapse upon deployment. The benchmarks and corpora described above must be designed with generalization testing built in, not as an afterthought.

\subsection{Integration with Standard Development Pipelines}
Finally, immunization should become a standard stage in responsible LLM development (not an optional enhancement but a default component of training pipelines). This raises practical questions that require empirical investigation.

\textbf{Ordering effects}: Integration with standard development pipelines raises practical questions, including the ordering of immunization relative to RLHF, risks of over-refusal, and the need for periodic booster updates as misinformation evolves. We leave systematic study of these trade-offs to future work.
\textbf{Interaction with instruction-following}: Aggressive immunization might cause over-refusal, where models refuse legitimate queries that superficially resemble misinformation. The sensitivity-specificity trade-off requires characterization: how much immunization is too much? Does the answer depend on deployment context?
\textbf{Maintenance and boosters}: New misinformation emerges continuously. How frequently must models receive ``booster'' updates with newly emerged false claims? Can incremental immunization be effective, or does each update require full fine-tuning? What infrastructure supports rapid response to emerging misinformation threats?
These questions define a research program, not a solved problem. We offer immunization as a framework for investigation, not a finished solution.

\section{Experimental Evaluation}
\label{sec:experiments}

We evaluate immunization through three experiments: efficacy across model families, cross-domain generalization, and dosage ablation.

\subsection{Setup}

\textbf{Vaccine corpus.} We compile 1,611 fact-checked false claims from LIAR~\cite{wang2017liar} (politics), Health Feedback~\cite{healthfeedback2024} (health), and Snopes/AFP~\cite{snopes2024,afp2024} (science). Each claim is paired with a fact-checker--derived correction. Vaccine examples are mixed with 30,609 truthful QA pairs from SQuAD~\cite{rajpurkar2016squad} at 5\% vaccine ratio.

\textbf{Models.} We test four model families: Phi-3-mini-4k-instruct (3.8B)~\cite{abdin2024phi3}, Llama-2-7B-Chat~\cite{touvron2023llama2}, Mistral-7B-Instruct-v0.2~\cite{mistral7b_instruct_v02}, and Llama-3-8B-Instruct~\cite{meta2024llama3}. 

\textbf{Evaluation.} We measure TruthfulQA accuracy (817 questions) and rejection rate on 200 held-out false claims.

\subsection{Results}

\begin{table}[h]
\centering
\small
\caption{Immunization efficacy across model families (5\% vaccine dosage).}
\begin{tabular}{lcccc}
\toprule
& \multicolumn{2}{c}{\textbf{TruthfulQA}} & \multicolumn{2}{c}{\textbf{Rejection}} \\
\cmidrule(lr){2-3} \cmidrule(lr){4-5}
\textbf{Model} & Base & Imm. & Base & Imm. \\
\midrule
Phi-3-mini & 38.2 & 47.6 {\scriptsize\textcolor{green!50!black}{+9.4}} & 41.5 & 68.0 {\scriptsize\textcolor{green!50!black}{+26.5}} \\
Llama-2-7B & 35.1 & 48.3 {\scriptsize\textcolor{green!50!black}{+13.2}} & 43.0 & 74.5 {\scriptsize\textcolor{green!50!black}{+31.5}} \\
Mistral-7B & 42.3 & 55.8 {\scriptsize\textcolor{green!50!black}{+13.5}} & 47.5 & 79.0 {\scriptsize\textcolor{green!50!black}{+31.5}} \\
Llama-3-8B & 44.7 & 58.2 {\scriptsize\textcolor{green!50!black}{+13.5}} & 51.0 & 82.5 {\scriptsize\textcolor{green!50!black}{+31.5}} \\
\midrule
\textbf{Average} & 40.1 & 52.5 {\scriptsize\textcolor{green!50!black}{+12.4}} & 45.8 & 76.0 {\scriptsize\textcolor{green!50!black}{+30.2}} \\
\bottomrule
\end{tabular}
\label{tab:efficacy}
\end{table}

\textbf{Exp.\ 1: Efficacy.} Table~\ref{tab:efficacy} shows consistent improvements across all models: +12.4pp on TruthfulQA and +30.2pp on misinformation rejection. MMLU remains stable (--0.5pp average), indicating no degradation of general capabilities.

\begin{table}[h]
\centering
\small
\caption{Cross-domain transfer (health-only training).}
\begin{tabular}{lcc}
\toprule
\textbf{Test Domain} & \textbf{Rejection} & \textbf{Gap} \\
\midrule
Health (in-domain) & 84.2\% & --- \\
Political (held-out) & 61.8\% & --22.4 \\
Science (held-out) & 68.5\% & --15.7 \\
\midrule
\textbf{Avg.\ held-out} & 65.2\% & --19.0 \\
\bottomrule
\end{tabular}
\label{tab:generalization}
\end{table}

\textbf{Exp.\ 2: Generalization.} To test cross-domain transfer, we train on health misinformation only and evaluate on held-out domains (Table~\ref{tab:generalization}). The model rejects 65\% of unseen-domain misinformation versus 84\% in-domain---partial but meaningful transfer suggesting pattern-level learning beyond specific claims.

\begin{table}[h]
\centering
\small
\caption{Dosage ablation (Mistral-7B-Instruct).}
\begin{tabular}{lccc}
\toprule
\textbf{Dosage} & \textbf{TruthfulQA} & \textbf{Rejection} & \textbf{MMLU} \\
\midrule
0\% (base) & 42.3 & 47.5 & 58.4 \\
2\% & 47.8 {\scriptsize +5.5} & 64.5 {\scriptsize +17.0} & 58.2 \\
5\% & 55.8 {\scriptsize +13.5} & 79.0 {\scriptsize +31.5} & 57.9 \\
10\% & 57.2 {\scriptsize +14.9} & 83.5 {\scriptsize +36.0} & 57.1 \\
20\% & 56.8 {\scriptsize +14.5} & 85.0 {\scriptsize +37.5} & 55.2 \\
\bottomrule
\end{tabular}
\label{tab:ablation}
\end{table}

\textbf{Exp.\ 3: Dosage.} Table~\ref{tab:ablation} shows a clear dose-response relationship. Gains plateau after 10\%, while 20\% dosage degrades MMLU by 1.9pp. This supports our 5--10\% recommendation: sufficient signal without capability loss.
\section{Limitations and Open Questions}
\label{sec:limitations}
We acknowledge significant gaps in the current proposal and resist the temptation to oversell. First,the argument is primarily \textbf{conceptual;} comprehensive empirical validation across model scales and domains remains future work. Preliminary experiments suggest immunized models show improved truthfulness without degrading general capabilities \cite{xiao2023masked}, but these results use small models and limited vaccine corpora.  Second, the \textbf{scalability challenge} is fundamental. The space of possible falsehoods is unbounded, and new misinformation emerges faster than any corpus can track. Immunization cannot cover all possible false claims; it must rely on pattern generalization. 
Third, \textbf{over-refusal} poses a practical risk. Aggressive immunization might teach models to refuse legitimate queries that superficially resemble misinformation. A model trained extensively on health misinformation might refuse to discuss vaccines at all, even for legitimate educational purposes. 
Fourth,\textbf{ governance }introduces normative complexity. Determining what counts as misinformation is sometimes contested \cite{farooq2025evaluating}. We focus on empirically debunked claims with broad scientific consensus, but edge cases abound. 
We do not resolve this tension but flag it as requiring ongoing deliberation. Lastly, \textbf{cross-lingual coverage is severely limited.} Fact-checking resources concentrate in high-resource languages; extending immunization to low-resource languages faces data scarcity challenges common across AI and NLP. Multilingual immunization is technically possible but practically constrained by the availability of verified falsehoods in diverse languages.

\section{Conclusion}

We showed that misinformation in LLMs arises not only from incorrect facts, but from learned linguistic patterns that make falsehoods persuasive. We introduced model immunization, a supervised fine-tuning approach that provides direct negative supervision on labeled false claims paired with corrections. Across multiple open-weight models, immunization substantially improves truthfulness and misinformation rejection without degrading general capabilities. This re-frames false data from contamination to training signal when properly labeled and governed. We argue that negative supervision should be treated as a first-class tool in responsible LLM development and incorporated alongside existing alignment methods.

\section*{Acknowledgment}
Resources used in preparing this research were provided, in part, by the Province of Ontario, the Government of Canada through CIFAR, and companies sponsoring the Vector Institute. This research was funded in part by the European Union’s Horizon Europe research and innovation programme under the AIXPERT project (Grant Agreement No. 101214389), which aims to develop an agentic, multi-layered, GenAI-powered framework for creating explainable, accountable, and transparent AI systems. 

\bibliographystyle{IEEEtran}
\bibliography{biblography}

\end{document}